\documentclass[letterpaper, 10 pt, conference]{ieeeconf}
\IEEEoverridecommandlockouts
\let\labelindent\relax
\usepackage{enumitem}
\usepackage{balance}
\usepackage[font={small}]{caption}
\usepackage[labelformat=simple]{subfig}
\usepackage{array}
\usepackage{textcomp}
\usepackage{mathtools, nccmath}
\usepackage{graphicx}
\usepackage{amsfonts}
\usepackage{amsmath}
\usepackage{amssymb}
\usepackage{algorithm}
\usepackage{algorithmic}
\usepackage{tikz}
\usepackage{arydshln}
\usepackage{multirow}
\usepackage{bm}
\usepackage{epstopdf}
\usepackage{siunitx}
\usepackage{xcolor}
\makeatletter
\let\NAT@parse\undefined
\makeatother
\usepackage[colorlinks=true,
            linkcolor=blue,
           ]{hyperref}

\newcommand{\figref}[1]{Fig.~\ref{#1}}
\newcommand{\secref}[1]{Section~\ref{#1}}

\title{\LARGE \bf Robotic Autonomous Trolley Collection with Progressive Perception and Nonlinear Model Predictive Control}
\author{Anxing Xiao$^\dagger$, Hao Luan$^\dagger$, Ziqi Zhao$^\dagger$, Yue Hong, Jieting Zhao, Weinan Chen, \\ Jiankun~Wang$^*$,~\IEEEmembership{Member,~IEEE}, Max~Q.-H.~Meng$^*$,~\IEEEmembership{Fellow,~IEEE}
\thanks{$^\dagger$ indicates equal contribution.}
\thanks{$^*$Corresponding authors: Jiankun Wang, Max Q.-H. Meng.}
\thanks{This work is partially supported by Shenzhen Key Laboratory of Robotics Perception and Intelligence (ZDSYS20200810171800001), Southern University of Science and Technology, and National Natural Science Foundation of China grant \#62103181.}
\thanks{All authors are with Shenzhen Key Laboratory of Robotics Perception and Intelligence, and the Department of Electronic and Electrical Engineering of Southern University of Science and Technology in Shenzhen, China. Max Q.-H. Meng is on leave from the Department of Electronic Engineering, the Chinese University of Hong Kong, Hong Kong, and also with the Shenzhen Research Institute of the Chinese University of Hong Kong, Shenzhen, China.
{\tt\small \{xiaoax@mail., luanh@mail., 12031215@mail., 12032838@mail., 12132162@mail., wangjk@, chenwn@\}sustech.edu.cn, max.meng@ieee.org.}
}
}

\begin{document}
\maketitle

\begin{abstract}
 
Autonomous mobile manipulation robots that can collect trolleys are widely used to liberate human resources and fight epidemics. 
Most prior robotic trolley collection solutions only detect trolleys with 2D poses or are merely based on specific marks and lack the formal design of planning algorithms.
In this paper, we present a novel mobile manipulation system with applications in luggage trolley collection. 
The proposed system integrates a compact hardware design and a progressive perception and planning framework, enabling the system to efficiently and robustly collect trolleys in dynamic and complex environments.
For perception, we first develop a 3D trolley detection method that combines object detection and keypoint estimation. 
Then, a docking process in a short distance is achieved with an accurate point cloud plane detection method and a novel manipulator design. 
On the planning side, we formulate the robot's motion planning under a nonlinear model predictive control framework with control barrier functions to improve obstacle avoidance capabilities while maintaining the target in the sensors' field of view at close distances. 
We demonstrate our design and framework by deploying the system on actual trolley collection tasks, and their effectiveness and robustness are experimentally validated. 
(Video\footnote{Video demonstration: \url{https://youtu.be/6SwjgGvRtno}.})

\end{abstract}

\section{Introduction}
\label{sec:Introduction}
    Robots have become popular in people's lives because they can complete tedious and complex tasks autonomously or collaboratively.
    In this paper, we discuss a robotic autonomous trolley collection system designed for trolley collection at airports.
    
    At airports, passengers usually use luggage trolleys to help carry luggage between gates and arrival/departure.
    For example, Hong Kong International Airport (HKG) has an annual passenger flow of more than 72.9 million passengers and has approximately 13,000 luggage trolleys\cite{wang_RealTimeDecisionMaking_2021_ITSMCS}.
    Naturally, for the convenience of passengers, collecting and redistributing these trolleys has become a vital but laborious task.
    Most airports, including HKG, still require considerable human resources to collect trolleys and return them to designated locations for continued use.
    However, the labor costs incurred by such human-driven operations are huge and continue to grow, especially in developed regions.
    Therefore, the robotic autonomous trolley collection system provides a promising and cost-effective solution for tedious and expensive tasks.
    
    In addition, with the outbreak of COVID-19, large international airports such as HKG have become high-risk areas for the spread of the virus.
    Airport staff working with trolleys are at risk of infection because the coronavirus can last for several days, even on inanimate objects including trolleys.
    Therefore, without human contact and intervention, robotic autonomous trolley collection will be another effort to break the chain of virus transmission as the pandemic escalates.
    In this paper, we focus on developing a robotic autonomous trolley collection system that integrates mechanical design, perception and planning, with the ability to navigate and reliably collect trolleys in complex environments.
    
    \begin{figure}[t]
        \centering
        \includegraphics[width=.98\linewidth]{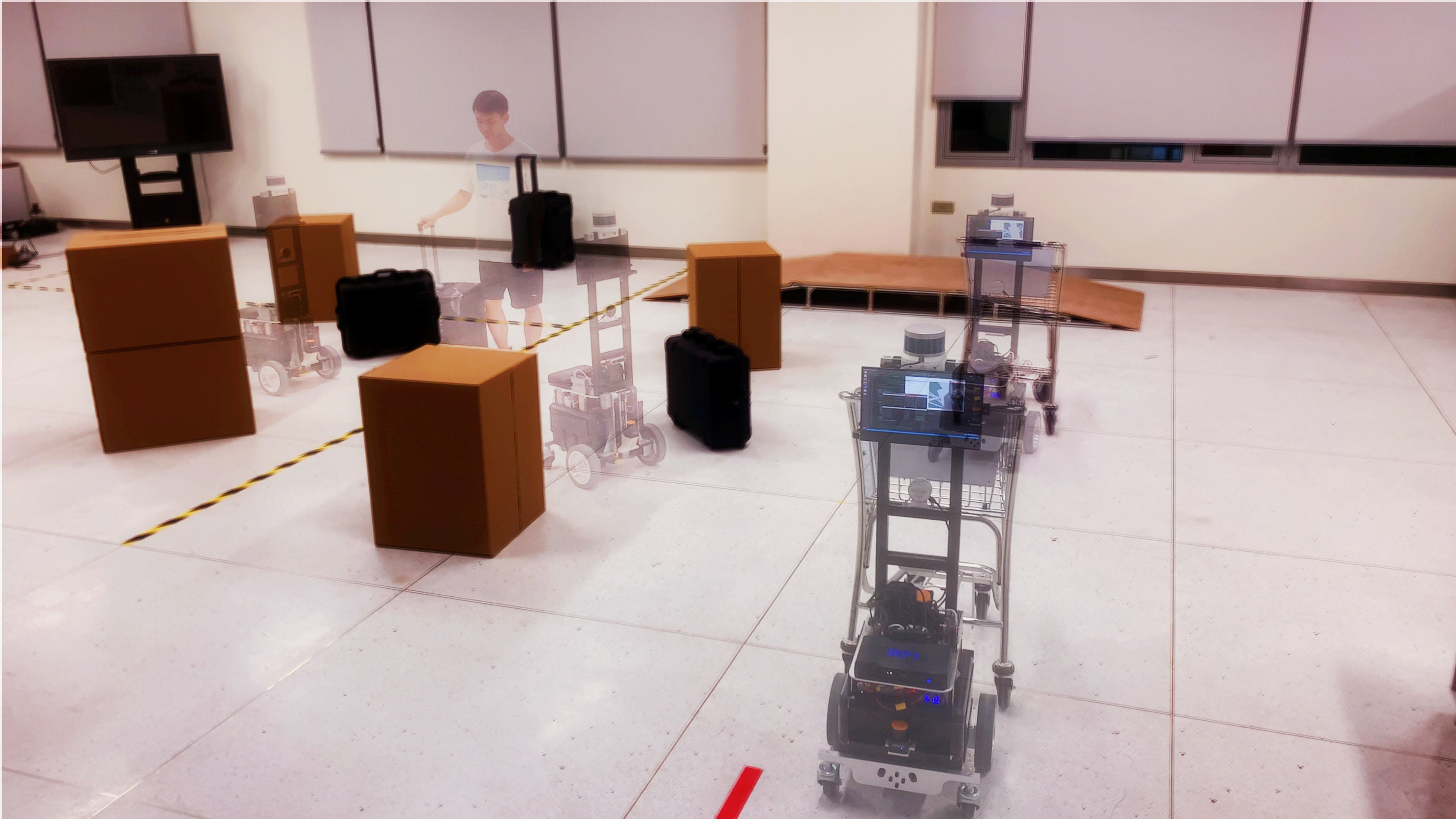}
        \caption{
            An autonomous robot detecting a trolley, safely navigating itself among people and obstacles, and collecting the trolley and transporting it to a designated location.
        }
        \label{fig:robot_trolley}
        \vspace{-0.3cm}
    \end{figure}

\subsection{Related Work}
\label{sub:related}
    An autonomous mobile manipulation robot that can find and manipulate objects, as shown in \figref{fig:robot_trolley}, is an engineering challenge featuring sophisticated incorporation of multiple modules, including mechanical design, perception, planning, and control. 
    The design of mobile manipulation platforms has been an active research area. 
    A few researchers take the approach of equipping mobile robots with robot arms, represented by the DLR-HIT-Hand\cite{4598864}, PR2\cite{bohren2011towards} and TIAGo\cite{pages2016tiago}. 
    However, these platforms are designed for a universal purpose using their sophisticated manipulators, so they lack reliability when performing repetitive tasks, e.g., collecting trolleys at airports. 
    The first robotic trolley collection solution is introduced in \cite{wang_CoarsetoFineVisualObject_2021_ISJ}. 
    The developed prototype has several sensors and a fork manipulator to lift a trolley. 
    Nonetheless, the lifting process is open-loop since there is no feedback sensor in the manipulator. 
    In our new design, we add sensors to introduce feedback detecting sudden impacts encountered by the manipulator. 
   
    For visual perception, learning-based 2D object detection such as Fast R-CNN\cite{ren2015faster} and YoloV5\cite{glenn_jocher_2021_4679653} has been well investigated in recent years. 
    These real-time object detection models endow mobile robots with the ability to localize specific targets in complex environments.
    In \cite{wang_CoarsetoFineVisualObject_2021_ISJ}, the authors use a trained R-CNN model to detect a trolley, and the robot moves towards the trolley while maintaining a bounding box of the target in the middle of the image.
    For accurate manipulation tasks, however, merely perceiving 2D information of the target is not enough. 
    The method in \cite{lin2019monocular} relies on markers pasted on the trolley despite its exploration in monocular 3D trolley detection.
    In autonomous driving, many researchers attempt to fully explore the potential of RGB images for 3D detection by recovering 3D objects from key points\cite{li2020rtm3d,he2019mono3d++}. 
    A key drawback of such methods is that when the target is too close to the camera, the limitation of the camera's field of view will cause failure in detection.
    In our method, to address the above shortcomings, we incorporate 2D detection and key point detection to estimate the 3D pose of a trolley at a long distance and leverage LiDARs to detect the backplane of the trolley at a short distance.
   
    For planning and control, there are some previous efforts at the trolley collection task. 
    In \cite{pan_SearchingSpaceConstrained_2021_ISJ,wang_CoarsetoFineVisualObject_2021_ISJ}, the adopted method is visual servoing. 
    The main disadvantage of such method is that it does not consider obstacle avoidance and safety, leaving another critical task on the to-do list upon field deployment. 
    The work of \cite{wang_PathPlanningNonholonomic_2020_2IICIRSI,wang_RealTimeDecisionMaking_2021_ITSMCS} focuses on task assignment and smooth-path generation for multiple robots with nonholonomic constraints, but safety is not a general consideration in the planning framework and the final docking error is not taken into account.
    Recently, optimization-based planning strategies such as model predictive control (MPC) have gained their prevalence in mobile robot planning and control\cite{yu2018mpc}. 
    There are also breakthroughs on tackling real-time safety guarantees for MPC with control barrier functions (CBFs)\cite{zeng_SafetyCriticalModelPredictive_2021_2ACCA,zeng_EnhancingFeasibilitySafety_2021_ACEMa,he2021lane-change-cbf}.
    CBFs are useful tools for integrating safety considerations as constraints into the general MPC framework.
    In our work, we formulate our motion planning problem under an MPC framework with obstacle avoidance constraints and field-of-view constraints, and it is validated more efficient and robust than the state of the arts concerning robotic autonomous trolley collection.


\subsection{Contributions}
    This work offers the following contributions: 
    \begin{enumerate}
        \item A novel robotic autonomous trolley collection system integrating a mechanical system and an efficient autonomy framework.
        \item A progressive perception strategy involving long-distance keypoints-based monocular 3D detection and short-distance accurate pose estimation using LiDARs.
        \item A safety-critical motion planner formulation under a nonlinear model predictive control framework with CBFs considering obstacle avoidance and field-of-view constraints. 
        \item Experimental demonstration in complex and dynamic environments of our system detecting target trolleys and safely collecting the trolleys.
    \end{enumerate}
    

\section{System Design}
\label{sec:framework}
The robots for collecting luggage trolleys in the airport need to replace trolley collectors to complete many tasks. 
Most of these tasks are very complex and challenging for a robotic system, so we specially designed a highly integrated hardware and software system suitable for trolley collection tasks in crowded environments.

\subsection{Mechanical System}
\label{subsec:mechanical}

\begin{figure}[h]
    \centering
    \includegraphics[width=1\linewidth]{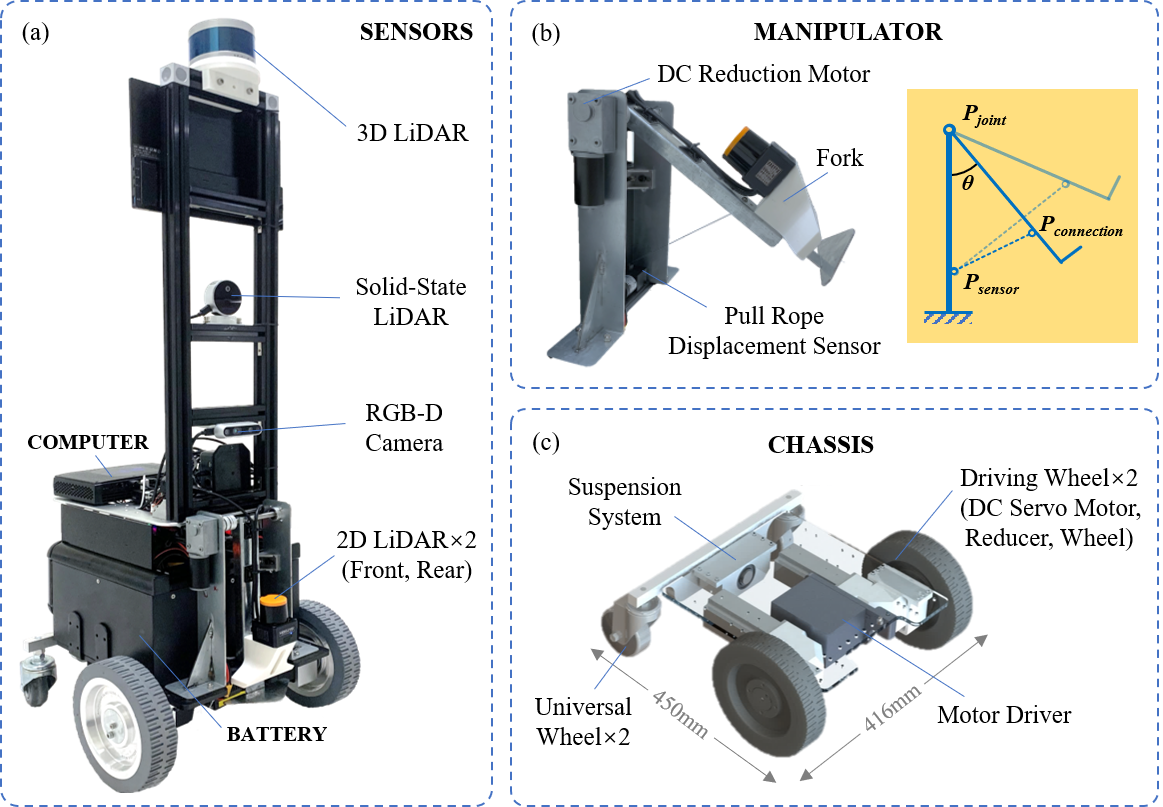}
    \caption{The robot consists of three main functional modules, namely, the chassis module (\figref{fig:design}(c)), the sensors module (\figref{fig:design}(a)), and the manipulator module (\figref{fig:design}(b)). In addition, the robot is equipped with a high-performance computer and a large-capacity battery to support stable operation (see \figref{fig:design}(a)).}
    \label{fig:design}
    \vspace{-0.3cm}
\end{figure}

The developed robot for luggage trolley collection, shown in \figref{fig:design}, is 1.2m high with 0.07m ground clearance, 0.45m long, and 0.416m wide.
As an integrated robotic system with mobile operation functions, the performance of movement, loading, and operation must be considered in the design.

\subsubsection{Chassis}
\label{subsec:chassis}

The two front wheels are driving wheels, shown in \figref{fig:design}(c).
Each driving wheel comprises a DC servo motor, a reducer, and a wheel, producing 31.75N.m of torque.
The rear wheels are two universal wheels connected to the car body by a suspension system. 

\subsubsection{Sensors}
\label{subsec:sensors}

For perception and localization, the robot is equipped with a 3D LiDAR, two 2D LiDARs, a solid-state LiDAR, and an RGB-D camera, as shown in \figref{fig:design}(a).
Due to adequate battery performance and sufficient onboard computing power, further extension of sensors and other equipment can be installed as required.

\subsubsection{Manipulator}
\label{subsec:Manipulator}

The manipulator is particularly designed for catching a luggage trolley in airports, shown in \figref{fig:design}(b). 
It is installed at the front of our robot and consists of a support base, a fork, a draw-wire encoder
, and a DC reducer motor. 
The motor lifts the fork in a rotating manner around a pivot, and the draw-wire encoder
serves as a feedback source for calculating the position of the fork based on the length of the wire.

The length variation $\Delta{l}$ of the wire reflects the speed and state of the fork movement. The length of the wire $l$ can be used to judge whether the fork is close to the designated position, and $\Delta{l}$ can be used to judge whether the fork is blocked or has grasped the trolley. Hence, by periodically detecting $l$ from the draw-wire encoder and through differential calculation, we can construct a feedback control system for the manipulator. 

\subsection{Autonomy Framework}

\begin{figure}[H]
    \centering
    \includegraphics[width=1.0\linewidth]{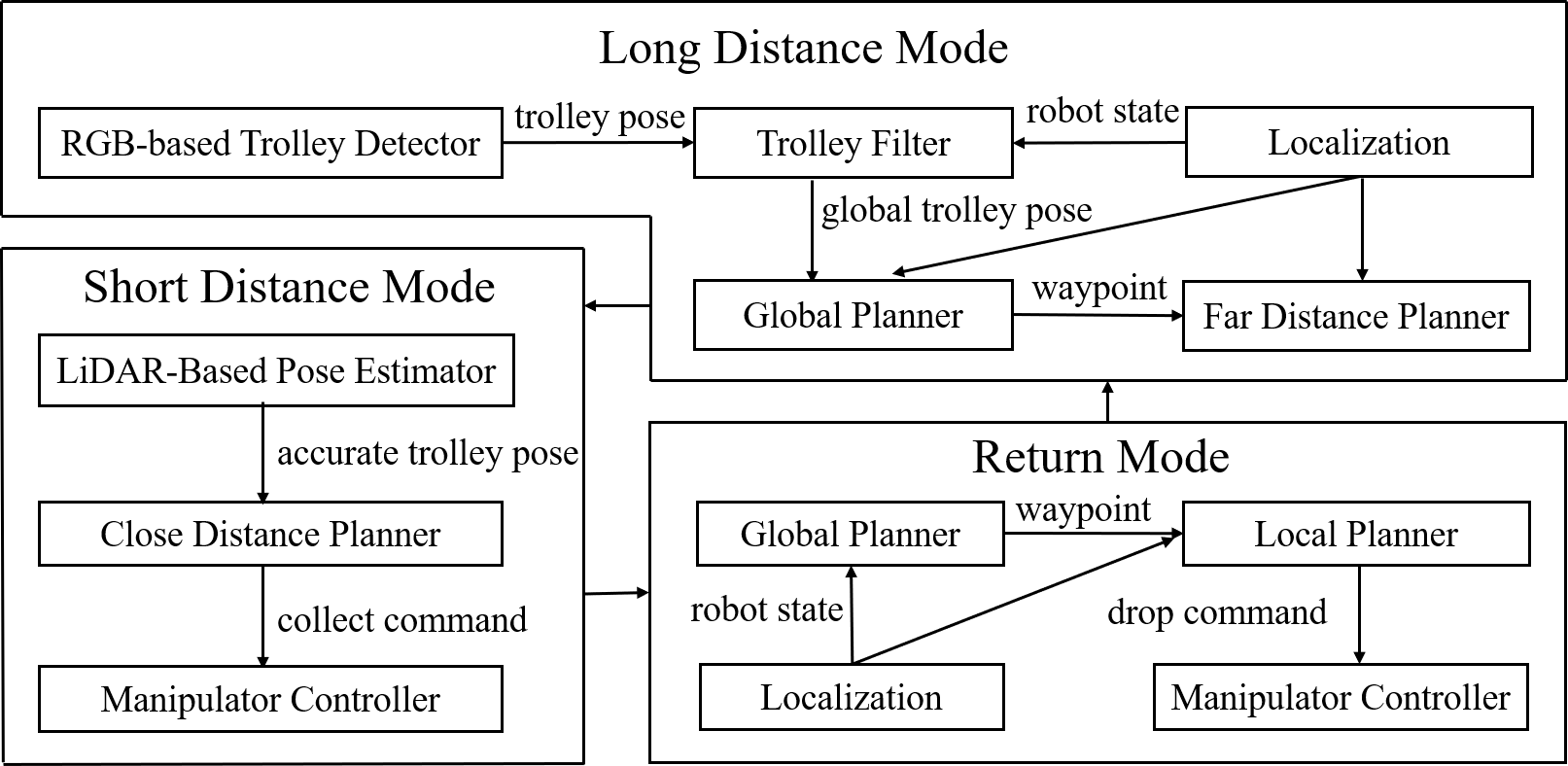}
    \caption{Autonomy framework overview.}
    \label{fig:framework}
    \vspace{-0.3cm}
\end{figure}

\figref{fig:framework} illustrates our navigation and collection autonomy. 
We propose a hierarchical framework to break the robotic autonomous trolley collecting process into three stages. 
\begin{figure}[htbp]
    \centering
    \subfloat[approaching stage]{\includegraphics[width=.34\linewidth]{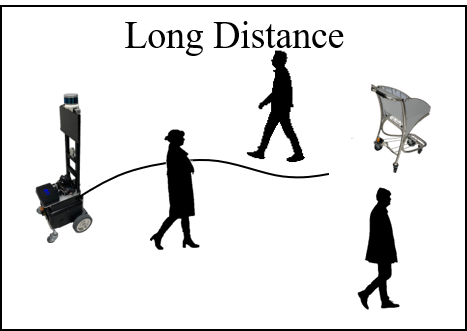}
    \label{subfig:approaching}}
    \subfloat[docking stage]{\includegraphics[width=.33\linewidth]{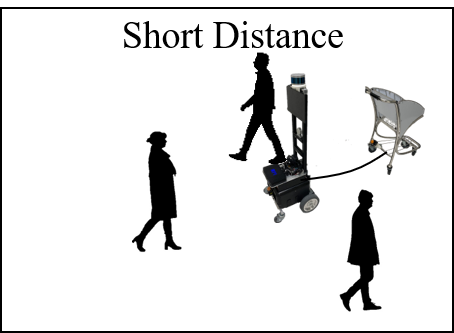}
    \label{subfig:docking}}
    \subfloat[returning stage]{\includegraphics[width=.33\linewidth]{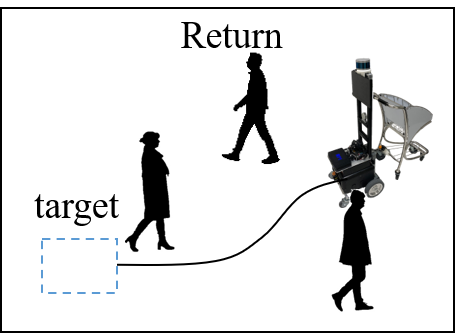}
    \label{subfig:return}}
    \caption{Illustration of the three stages of our framework.
        (a) At the approaching stage, the robot detects the trolley at long distances and navigates safely in crowded environments.
        (b) At the docking stage, the robot catches the trolley with fine motions.
        (c) At the returning stage, the robot transports the trolley to a returning spot.
    }
    \vspace{-0.4cm}
\end{figure}

In the approaching stage shown in \figref{subfig:approaching}, the robot moves in a crowded environment and finally gets to the back of a target trolley and shares the same orientation as the trolley. 
An RGB-D camera is used at this stage to perceive the trolley's position and orientation at a fairly long distance. 
The planner generates a motion trajectory to approach the trolley while avoiding obstacles in the dynamic environment.
In the docking stage (see \figref{subfig:docking}), the robot continues to move towards an ideal docking position precisely and then catches the trolley with its manipulator. 
At this stage, the robot should accurately get to the docking position so that the final aligning error can be small enough for a successful catch.
When the robot arrives at the exact docking location, the planner will give the low-level manipulator controller an action command to perform the final catch. 
After successfully capturing the target trolley, the robot carries the trolley to a designated returning spot, as \figref{subfig:return} shows. 
During all stages, an occupancy grid map is built with the Gmapping package\cite{gmapping}, and the AMCL\cite{amcl} localization is utilized to estimate the robot's states in the world frame. 

\section{Progressive Perception And Planning Strategy}
\label{sec:progressive}
In this section, we characterize our trolley collection strategy as a two-stage process based on the distance between the trolley and the robot. 
The collection task is simplified to a planar model since we assume that the trolley and the robot are in the same plane. 
At long distances, we utilize a monocular camera to detect the trolley and roughly estimate its three dimensional pose $\mathbf{q_{\text{tar}}} = [x_{\text{tar}}, y_{\text{tar}}, \theta_{\text{tar}}]^T$, wherein $\mathbf{p}_{\text{tar}} = [x_{\text{tar}}, y_{\text{tar}}]^T$ denotes the trolley's position and $\theta_{\text{tar}}$ represents its orientation. 
At short distances, the trolley's pose is precisely estimated by using a LiDAR.

\subsection{Monocular 3D Trolley Detection at Long Distance}

\begin{figure}[H]
    \centering
    \includegraphics[width=1.0\linewidth]{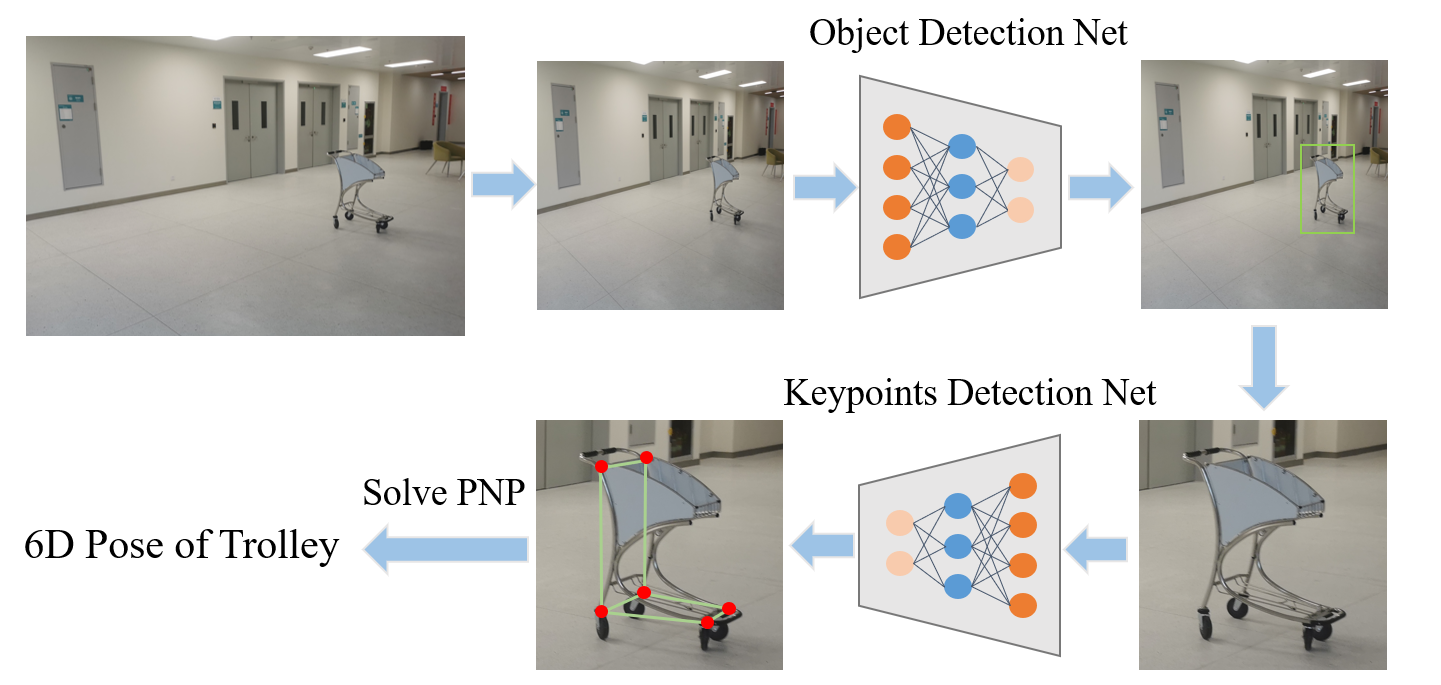}
    \caption{In the monocular 3D pose detection, a source image is first downsampled and put into an object detection net, then the detected trolley is cropped out from the original image and put into a keypoint detection net, and eventually the detected keypoints are used to solve a PnP problem.}
    \label{fig:rgbdetect}
    \vspace{-0.3cm}
\end{figure}

The monocular 3D detection framework consists of three parts as shown in \figref{fig:rgbdetect}. 
First, we use an object detection network to find a 2D bounding box of the target trolley from a downsampled RGB image $I_s \in \mathbb{R}^{W_s \times H_s \times 3}$. 
Then, from the original high-resolution image $I \in \mathbb{R}^{W \times H \times 3}$, we crop the trolley image with the bounding box and get a new image $I_c \in \mathbb{R}^{W_c \times H_c \times 3}$ . 
Second, instead of predicting the 3D pose of the trolley directly, we use a deep neural network to predict six 2D keypoints (red points in \figref{fig:rgbdetect}) in the cropped image $I_c$. 
Finally, with the \emph{a priori} known 3D coordinates of the 2D keypoints, we can calculate the corresponding 6D pose by minimizing the reprojection error of the keypoints in the original image $I$.

Specifically, in the first part, we choose YOLOV5\cite{glenn_jocher_2021_4679653} as our network model for real-time trolley detection due to its efficiency in object detection.
In the second part, inspired by the human pose estimation\cite{newell2016stacked}, we adopt the stacked hourglass network structure to estimate the heatmaps of six 2D keypoints $\hat{p}_i^c = [\hat{x}_{i}^c, \hat{y}_{i}^c]^T$ for $i = 0,1,\ldots,5$, in the cropped image $I_c$.
Then we can get the corresponding homogeneous 2D keypoints $\hat{p}_i = [\hat{u}_{i}, \hat{v}_{i}, 1]^T$ for $i = 0,1,\ldots,5$, in image coordinates of the original image $I$. 
In the third part, we solve a perspective-n-point (PnP) problem\cite{newell2016stacked} to obtain the pose of the trolley.
According to the perspective projection model for cameras, we have the following relationship
\begin{equation}
\label{eq:projection}
    s_i p_i  = KX_i^c =  K \begin{bmatrix} R \;|\; T \end{bmatrix} X_i^t , \quad i = 0, 1, \ldots, M
\end{equation}
where $X_i^c = [x_i^c,y_i^c,z_i^c,1]^T$ and $X_i^t =  [x_i^t,y_i^t,z_i^t,1]^T$ represent the homogeneous 3D coordinates of  the keypoints in the camera's frame and the trolley's frame, respectively; 
$K$ is the intrinsic camera matrix that projects $X_i^t$ to the image point $p_i = [u_i, v_i, 1]^T$ in homogeneous image coordinates; 
$s_i$ is a scale factor. 
Then, we solve the PnP problem to get a 3D rotation $R$ and a 3D translation $T$ from the trolley's frame to the camera's frame by adopting the EPnP algorithm\cite{lepetit2009epnp}. 
Eventually, with localization information of the robot base and the relative pose of the trolley in the camera frame, we can calculate the global state of the trolley $\mathbf{q_{\text{tar}}} = [x_{\text{tar}}, y_{\text{tar}}, \theta_{\text{tar}}]^T$. 
Moreover, a filter is performed to avoid sudden changes in detection results since the trolley should be static most of the time. 
If the trolley is not in the camera's field of view (FoV), the state of the trolley is set to be the same as the last time when it was within the FoV.

\subsection{LiDAR-Based Pose Estimation in Short Distance}

\begin{figure}[H]
    \centering
    \includegraphics[width=1.0\linewidth]{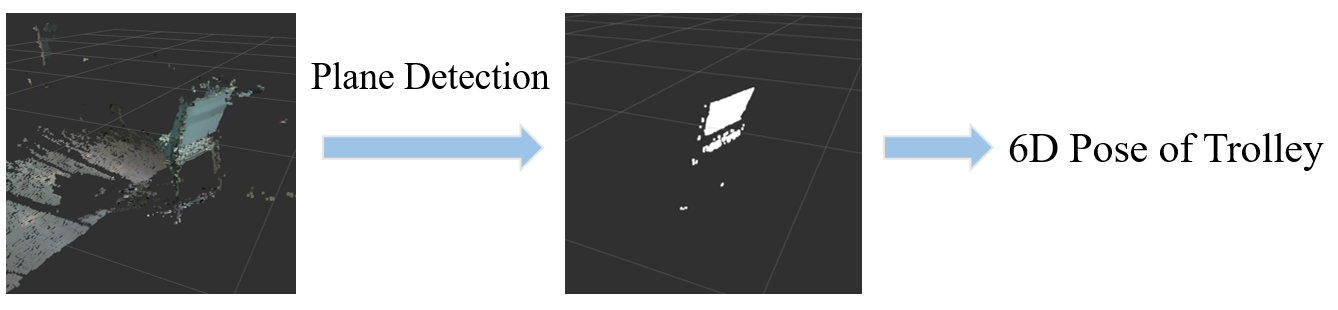}
    \caption{Accurate trolley pose estimation enabled by plane detection using a solid-state LiDAR.}
    \label{fig:lidardetect}
    \vspace{-0.3cm}
\end{figure}

During docking, accurate perception of the trolley's pose is vitally crucial. 
\figref{fig:lidardetect} illustrates our perception strategy at this stage. 
We use a solid-state LiDAR to yield a point cloud, and then perform plane detection and fitting with those points of the cloud to estimate the precise pose of the trolley. 
The obtained point cloud is noted by $ P = \{ p_1, p_2, \ldots, p_n \}$ where $ p_i \in \mathbb{R}^3 $ for $i=1,\ldots,n$, with $n$ being the total number of points. 
To get an ideal point cloud characterizing the backplane of a trolley, the robot should be at a suitable pose. 
Empirically, we set the robot facing the backplane close behind the trolley (0.3m$\sim$2m). 
Since the trolley's pose obtained at the approaching stage is with a decimeter-level precision, equipped with a good enough motion planner, which we will show in \secref{sub:motionplanning}, our LiDAR's FoV is large enough to ensure the backplane will be presented entirely in the point cloud.

To estimate the trolley's pose $\mathbf{q}_{\text{tar}}$ at a centimeter-level precision, we filter out interference points by setting a threshold.
After that, we conduct plane fitting through the RANSAC algorithm\cite{fischler1981random} and get a set of plane points $ P_\text{plane} = \{ p_1, p_2, ..., p_M \} \subseteq P$ and 4 parameters $a$, $b$, $c$, and $d$ in the plane equation $ aX + bY + cZ + d = 0 $. 
With the plane parameters and the plane points, we may estimate the center point and the yaw angle of the back plane by calculating the center point of the filtered point cloud and the normal vector of the plane. 
After obtaining the trolley's 3D pose $\mathbf{q}_{\text{tar}}$, we can then figure out a manipulation pose for the robot to collect the trolley.

\subsection{Safety-Critical Motion Planning with FOV Constraints}
\label{sub:motionplanning}
This planning part considers two main problems, videlicet,  generating a feasible state and control trajectory, and avoiding unsafe actions in crowded environments. 
In both stages of the collection process, the robot needs to move to a given goal state $\mathbf{x_{\text{goal}}} = [x_{\text{goal}}, y_{\text{goal}}, \theta_{\text{goal}}]^T$. 
Concretely, the goal state at the approaching stage is at a position behind the trolley and an orientation same as the trolley, while at the docking stage, the goal state is an ideal pose for the robot to operate manipulator. 

In this work, we characterize the safe set $\mathcal{C}$ of states as the zero-superlevel set of a continuously differentiable function $h: \mathcal{X} \subseteq \mathbb{R}^3 \rightarrow \mathbb{R}$ 
\begin{equation}
\label{eq:safeset}
    \mathcal{C} = \{ \mathbf{x} \in \mathcal{X} : h(\mathbf{x}) \geq 0  \} 
.\end{equation}
Safety, in our case, has a physical meaning of avoiding all static and dynamic obstacles. 
To do so, we can keep the distance between the robot and any obstacles beyond a specific range.
Therefore, it is natural to define the following function to construct our safe set $\mathcal{C}$
\begin{equation}
\label{eq:safeCBF}
    h_{\text{ob}}(\mathbf{x}) = (x-x_{\text{ob}})^2 +  (y-y_{\text{ob}})^2 - d_{\text{safe}}^2
\end{equation}
where $\mathbf{x}_{\text{ob}}=[x_{\text{ob}}, y_{\text{ob}}]^T$ denotes the position of any obstacle and $d_{\text{safe}}$ a predefined safety distance. 

\begin{figure}[H]
    \centering
    \includegraphics[width=0.45\linewidth]{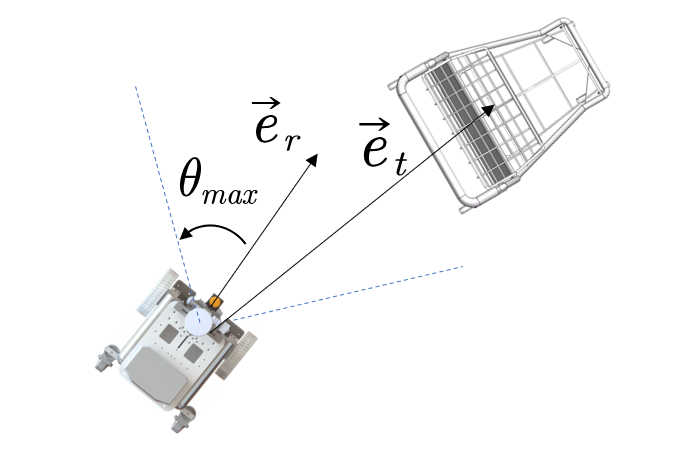}
    \caption{Illustration of the maintain-field-of-view requirement.}
    \label{fig:view}
    \vspace{-0.3cm}
\end{figure}

At the docking stage, it is preferable to let the trolley remain in the FoV of the solid-state LiDAR. 
As is shown in \figref{fig:view}, $\vec{\mathbf{e}}_t$ is a unit vector starting from the robot and pointing at the trolley; $\vec{\mathbf{e}}_r$ represents a unit vector in the direction of the robot's orientation; $\theta_{\text{max}}$ is the maximal angle between these two vectors at which the trolley stays within the LiDAR's FoV. 
To meet this requirement of maintaining observation, we define the following function that we will use later in our planning formulation: 
\begin{equation}
\label{eq:viewCBF}
    h_{\text{view}}(\mathbf{x}) = \vec{\mathbf{e}}_t \cdot \vec{\mathbf{e}}_r  - \cos{\theta_{\text{max}}}
.\end{equation}

Then, we introduce CBF constraints \cite{zeng_SafetyCriticalModelPredictive_2021_2ACCA,zeng_EnhancingFeasibilitySafety_2021_ACEMa}
\begin{equation}
\label{eq:constraints}
        \Delta h(\mathbf{x}_{k}, \mathbf{u}_{k})+\lambda_k h(\mathbf{x}_{k}) \ge 0
,\end{equation}
where $\Delta h(\mathbf{x}_{k},\mathbf{u}_{k}) := h(\mathbf{x}_{k+1}) - h(\mathbf{x}_{k})$ with $\lambda_k\in (0,1]$. 
This kind of constraints ensures $h$ becomes a discrete-time CBF, which means the safe set $\mathcal{C}$ defined in \eqref{eq:safeset} is invariant along the trajectories of a discrete-time dynamic system. 
Also, one can find that \eqref{eq:constraints} guarantees the lower bound of $h$ decreases exponentially at time $k$ with the rate $1-\lambda_k$. 

We formulate the planning task as a nonlinear model predictive control (NMPC) problem. 
At the approaching stage, the formulation has the following form: 
\begin{subequations}
\label{eq:optimization-formulation}
    \begin{align}
        \min_{ \{ \mathbf{x}_k, \mathbf{u}_k \} } \ & \left\| \mathbf{x}_N - \mathbf{x}_{\text{goal}} \right\|^2_{P_f}
    	+\sum_{k=0}^{N-1} {\left\| \mathbf{u_k} \right\|^2_{Q_u}
    	 } \label{subeq:opti} \\
        \text{s.t.} \ & \mathbf{x}_{k+1}=f(\mathbf{x}_k,\mathbf{u}_k) \label{subeq:dynamics} \\
        & \mathbf{x}_0= \mathbf{x}_{\text{init}} \label{subeq:initial-condition}\\
        & \mathbf{x}_k\in \mathcal{X} ,\mathbf{u}_k\in \mathcal{U} \label{subeq:admissible} \\
        & \Delta h_{\text{ob}}^{i}( \mathbf{x}_{k}, \mathbf{u}_{k})+\lambda_k h_{\text{ob}}^{i}( \mathbf{x}_k)\ge 0 \label{subeq:obconstraint}
    \end{align}
\end{subequations}
where $\|\mathbf{x}\|_A := \sqrt{\frac{1}{2}\mathbf{x}^T A \mathbf{x}}$, and the two positive definite matrices $P_f$ and $Q_u$ are respectively coefficients measuring terminal costs and running control costs. 
\eqref{subeq:opti} minimizes the quadratic cost function over a horizon of $N$ steps. In \eqref{subeq:dynamics}, we use the differential-drive model as the robot's system model.
\eqref{subeq:admissible} constrains the states and control inputs in a reachable state set and an admissible control set, respectively. 
Constraint \eqref{subeq:obconstraint} is for obstacle avoidance. 

At the docking stage, the formulation is similar:
\begin{subequations}
\label{eq:optimization-formulation-2}
    \begin{align}
        \min_{ \{ \mathbf{x}_k, \mathbf{u}_k, \delta_k \} } \ & \left\| \mathbf{x}_N - \mathbf{x}_{\text{goal}} \right\|^2_{P_f}
    	+\sum_{k=0}^{N-1} {\left\| \mathbf{u_k} \right\|^2_{Q_u} + w\delta_k^2
    	 } \label{subeq:opti2} \\
        \text{s.t.} \ & \mathbf{x}_{k+1}=f(\mathbf{x}_k,\mathbf{u}_k)  \\
        & \mathbf{x}_0= \mathbf{x}_{\text{init}} \\
        & \mathbf{x}_k\in \mathcal{X} ,\mathbf{u}_k\in \mathcal{U}  \\
        & \Delta h_{\text{ob}}^{i}( \mathbf{x}_{k}, \mathbf{u}_{k})+\lambda_k h_{\text{ob}}^{i}( \mathbf{x}_k)\ge 0 \label{subeq:obconstraint2} \\
        & \Delta h_{\text{view}}( \mathbf{x}_{k}, \mathbf{u}_{k})+\mu_k h_{\text{view}}( \mathbf{x}_k)\ge \delta_k \label{subeq:viewconstraint}
    \end{align}
\end{subequations}
At this stage, we describe the states at which the trolley remains in the robot's FoV as a safe set defined by joining \eqref{eq:safeset} and \eqref{eq:viewCBF}. 
Similar to the obstacle avoidance constraint \eqref{subeq:obconstraint2}, the maintaining observation requirement is formulated as the constraint \eqref{subeq:viewconstraint}. 
To avoid infeasibility, we introduce a slack variable $\delta$ and minimize it by the cost term $w \delta^2$. 

Upon implementation, these optimization problems are formulated in CasADi\cite{andersson2019casadi} and solved with IPOPT\cite{biegler2009large}.

\section{Implementation and Experiments}
\label{sec:results}


\subsection{Perception System Evaluation}
\label{sub:perceptionEvaluation}

\subsubsection{Data Set}
To ensure the robustness of the detection task at long distances, we built our own data set for network training. 
For object detection, we collected 1,200 pictures of the trolleys with different illumination, backgrounds, angles, etc. 
These pictures were all labeled with 2D bounding boxes around the trolleys.
The data set was divided into three parts, namely, 800 for training, 200 for validation, and 200 for testing. 
For key points detection, we prepared 800 pictures of the cropped trolleys images with accurate key point labels. 
Similarly, we arranged 600 images for training, 100 for validation, and another 100 for testing. 

\subsubsection{Implementation Details}
We implemented and trained our monocular 3D trolley detection networks offline using PyTorch on an Intel machine with an i7-9750H CPU and an NVIDIA GTX 1660Ti GPU. 
Our 2D detection network is adapted from the official code releases of YOLOV5 \cite{glenn_jocher_2021_4679653}. 
Training this network on our own data set, we adopted the SGD optimizer\cite{bottou2010large} for 300 epochs in total with a batch size of 16.
The base learning rate was $0.01$, and we reduced it to $0.001$ from the 150th epoch and to $0.0001$ from the 200th epoch.
The training stage of the object detection net lasted for roughly 14 hours. 
For key points detection, we used a stacked hourglass network with PyTorch upon implementation. 
To improve the generalization ability of our model, we leveraged data augmentation techniques such as random scaling, cropping, flipping, and color transformation.
During training, we run the Adam\cite{kingma2014adam} optimizer with a base learning rate of $0.0001$ for the first 200 epochs, and reduced it at a decreasing rate of $0.95$ every 10 epochs later on. 
Finally, it took about 4 hours to train our key points detection network with a batch size of 8. 
At short distances, our method based on plane detection does not involve learning, so we implemented it with the Point Cloud Library\cite{rusu20113d}.

\subsubsection{Perception Results Evaluation}

\begin{figure}[htb]
    \centering
    \includegraphics[width=1\linewidth]{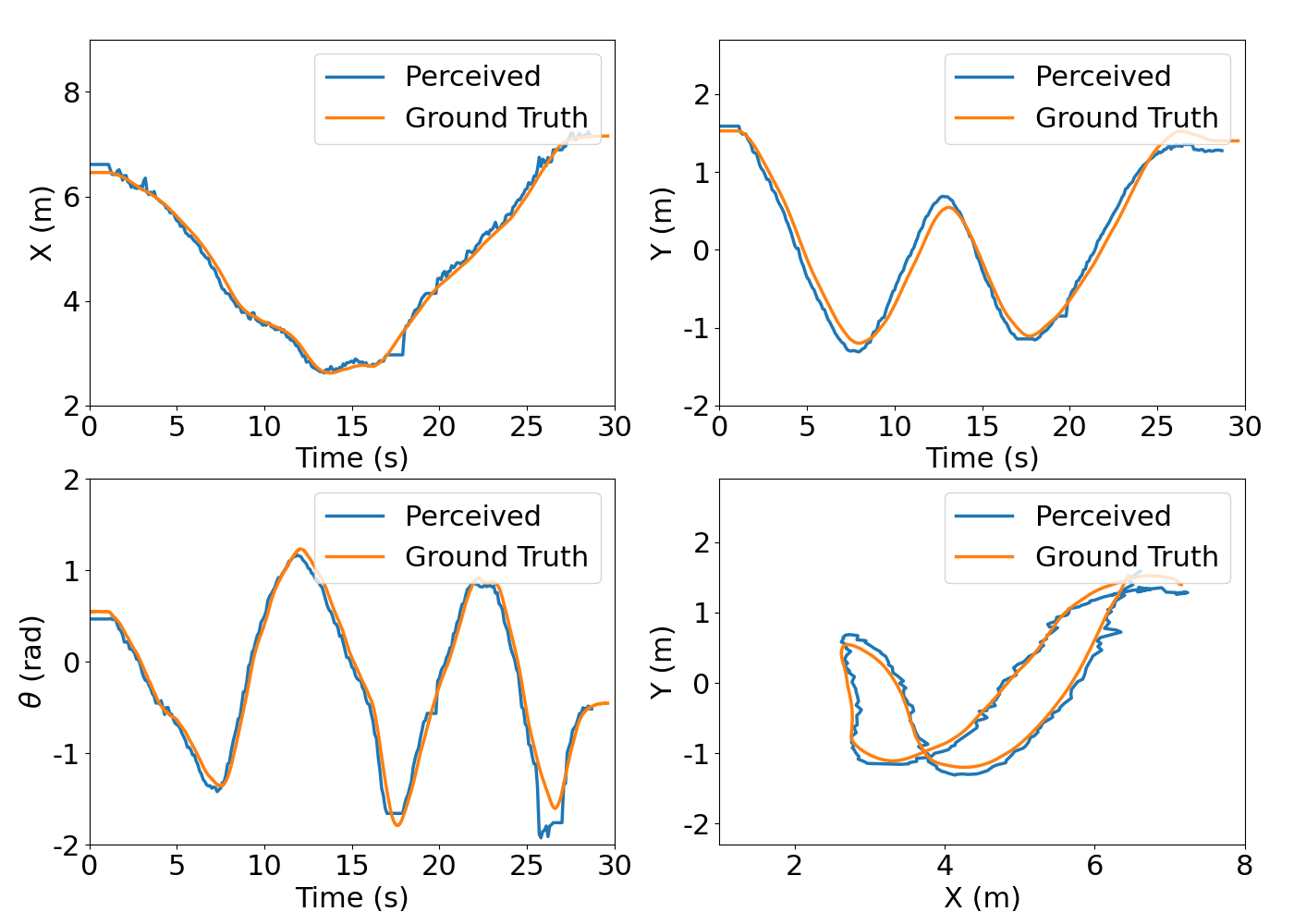}
    \caption{Comparisons between the ground truth pose of a moving trolley and results of our 3D monocular method. 
    These four subfigures represent the $x$ coordinate, $y$ coordinate, orientation $\theta$, and position trajectory respectively.}
    \label{fig:expPercption1}
    \vspace{-0.3cm}
\end{figure}

\begin{figure*}[!t]
    \centering
    \subfloat[safely approaching the target]{\includegraphics[width=.245\linewidth]{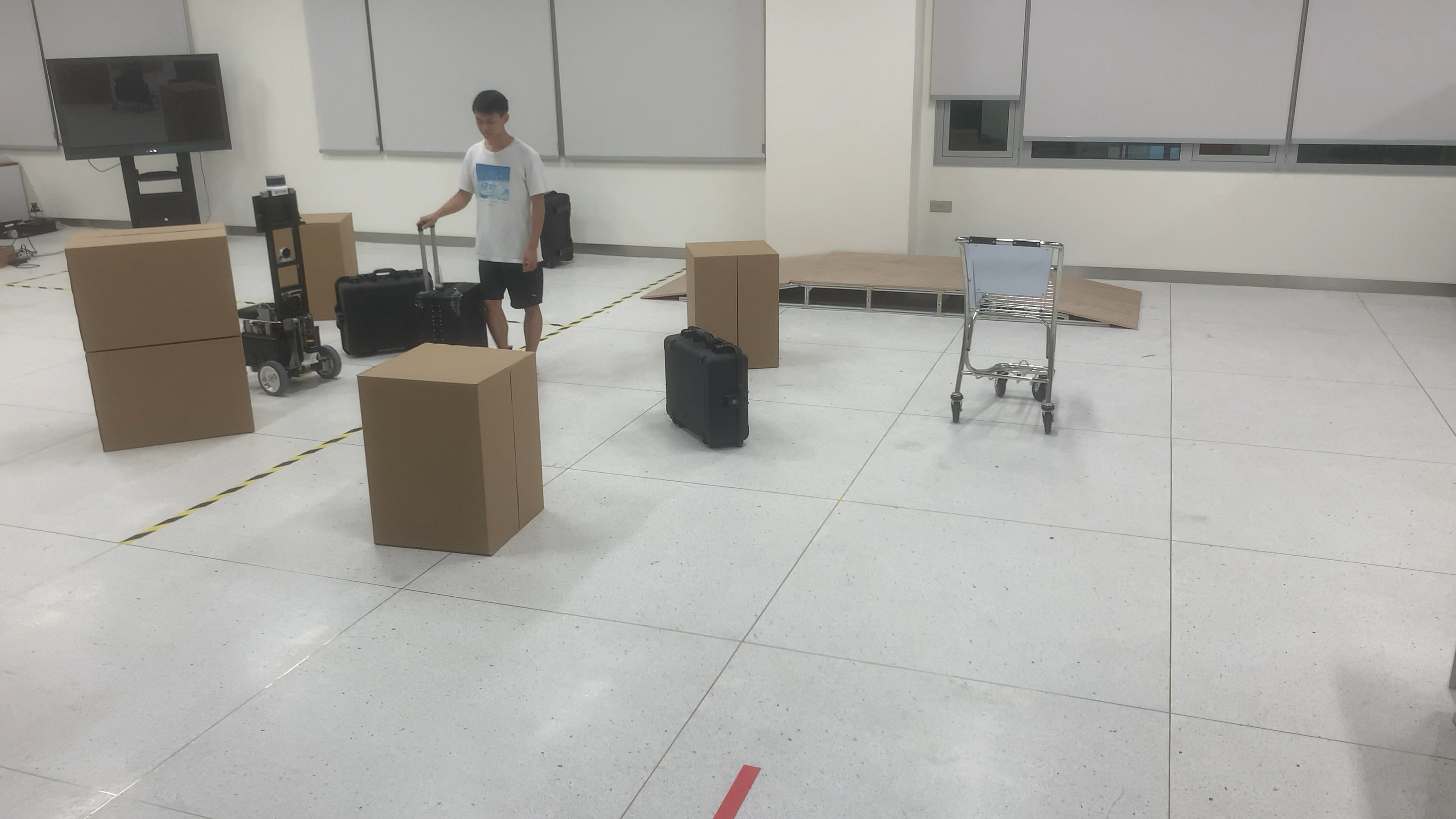}
    \label{subfig:demo1}}
    \subfloat[proceeding to the docking pose]{\includegraphics[width=.245\linewidth]{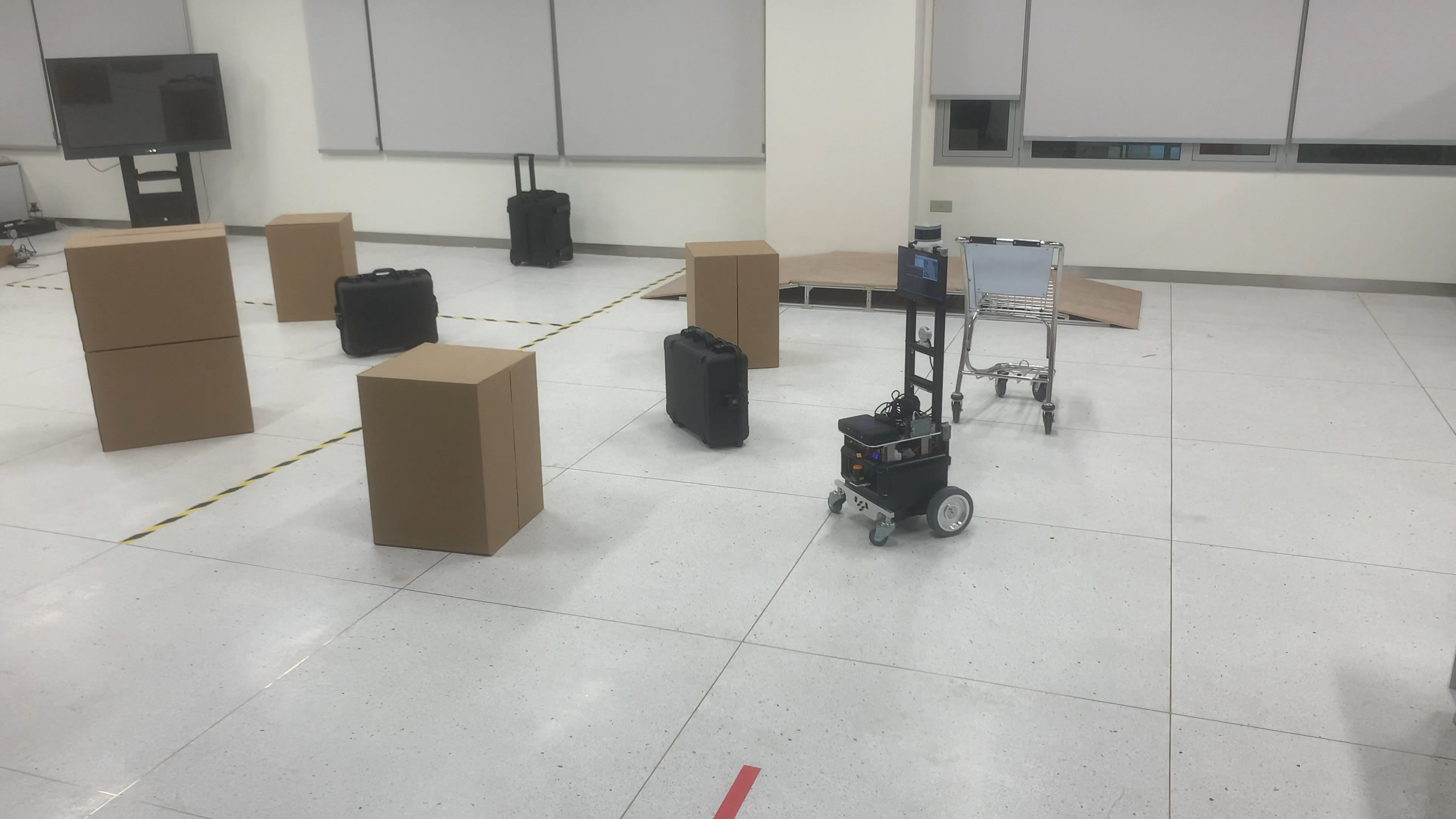}
    \label{subfig:demo2}}
    \subfloat[collecting the trolley]{\includegraphics[width=.245\linewidth]{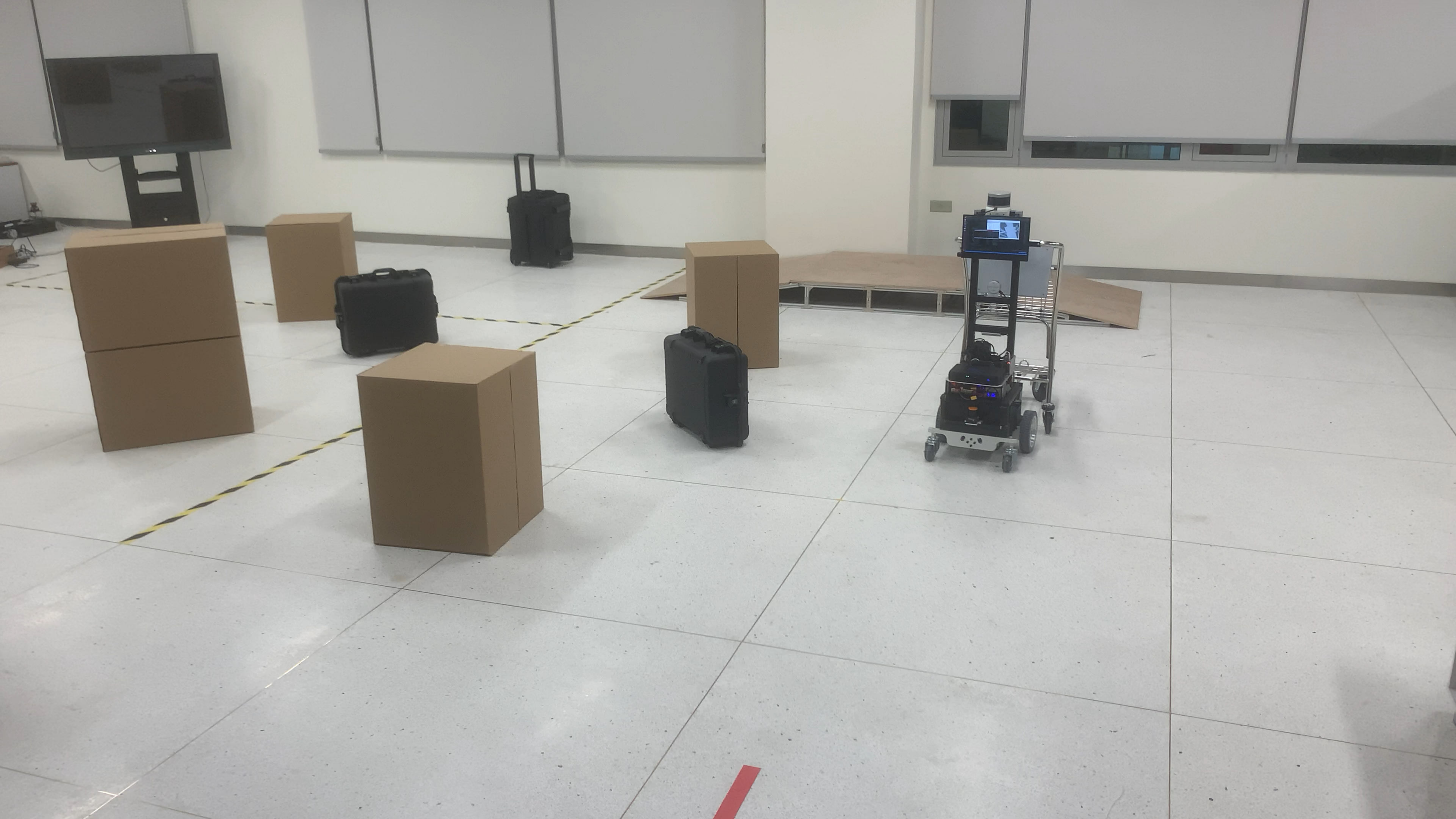}
    \label{subfig:demo3}}
    \subfloat[returning the trolley]{\includegraphics[width=.245\linewidth]{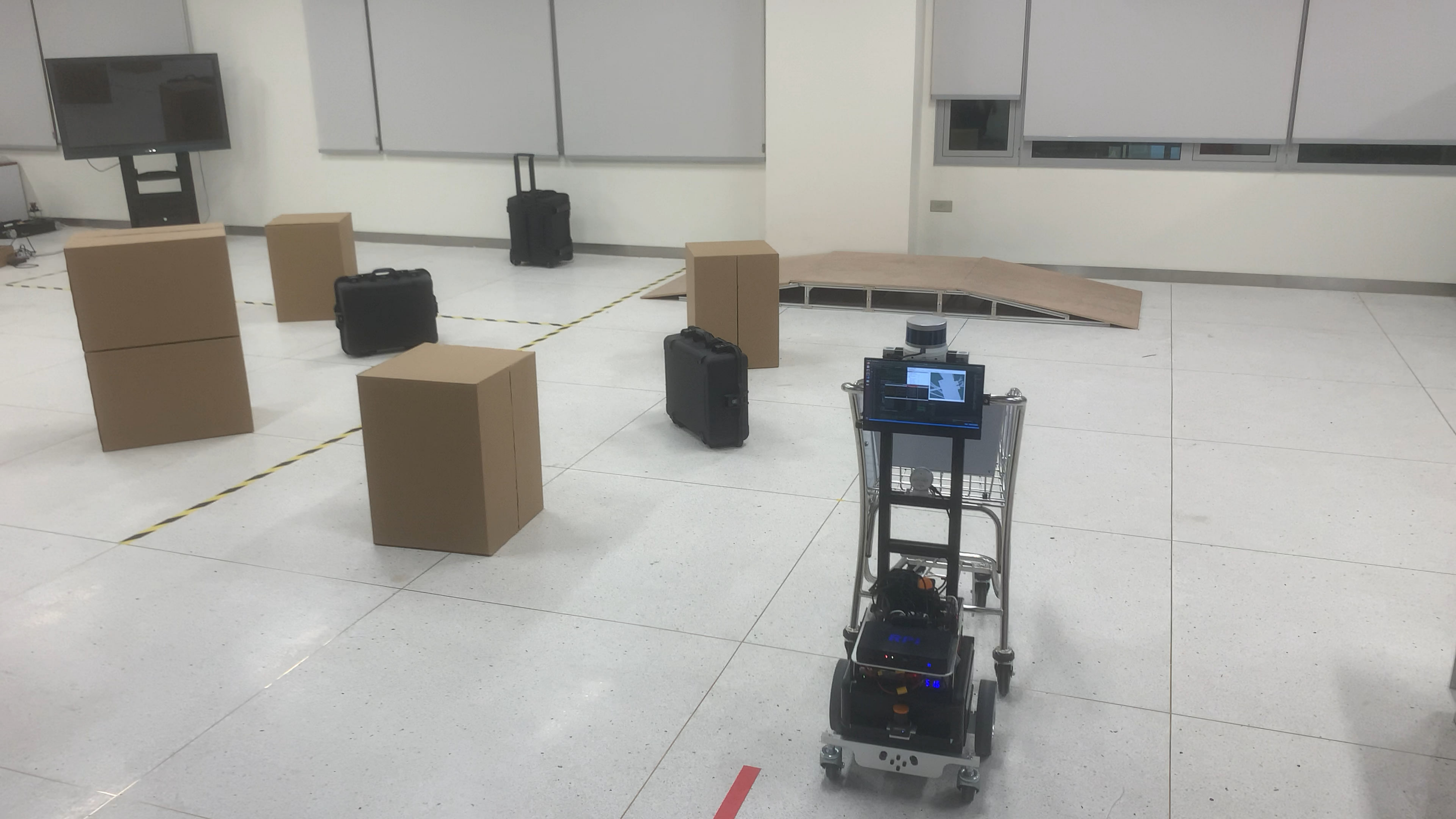}
    \label{subfig:demo4}}
    \caption{Snapshots of demonstration of our system conducting an actual trolley collection task. 
    (a) When the robot was approaching the trolley, a human with a suitcase moved across the robot's route right in front of it. 
    (b) The robot then slowed down, adjusted its route to avoid the human and other obstacles, and arrived at the goal position for docking. 
    (c) The robot reached the exact manipulation pose based on its own perception and planning in real time. 
    (d) After a successful capture, the robot carried the trolley to the returning spot. 
    }
    \label{fig:demo}
    \vspace{-0.5cm}
\end{figure*}

\begin{figure}[htbp]
    \centering
    \includegraphics[width=1\linewidth]{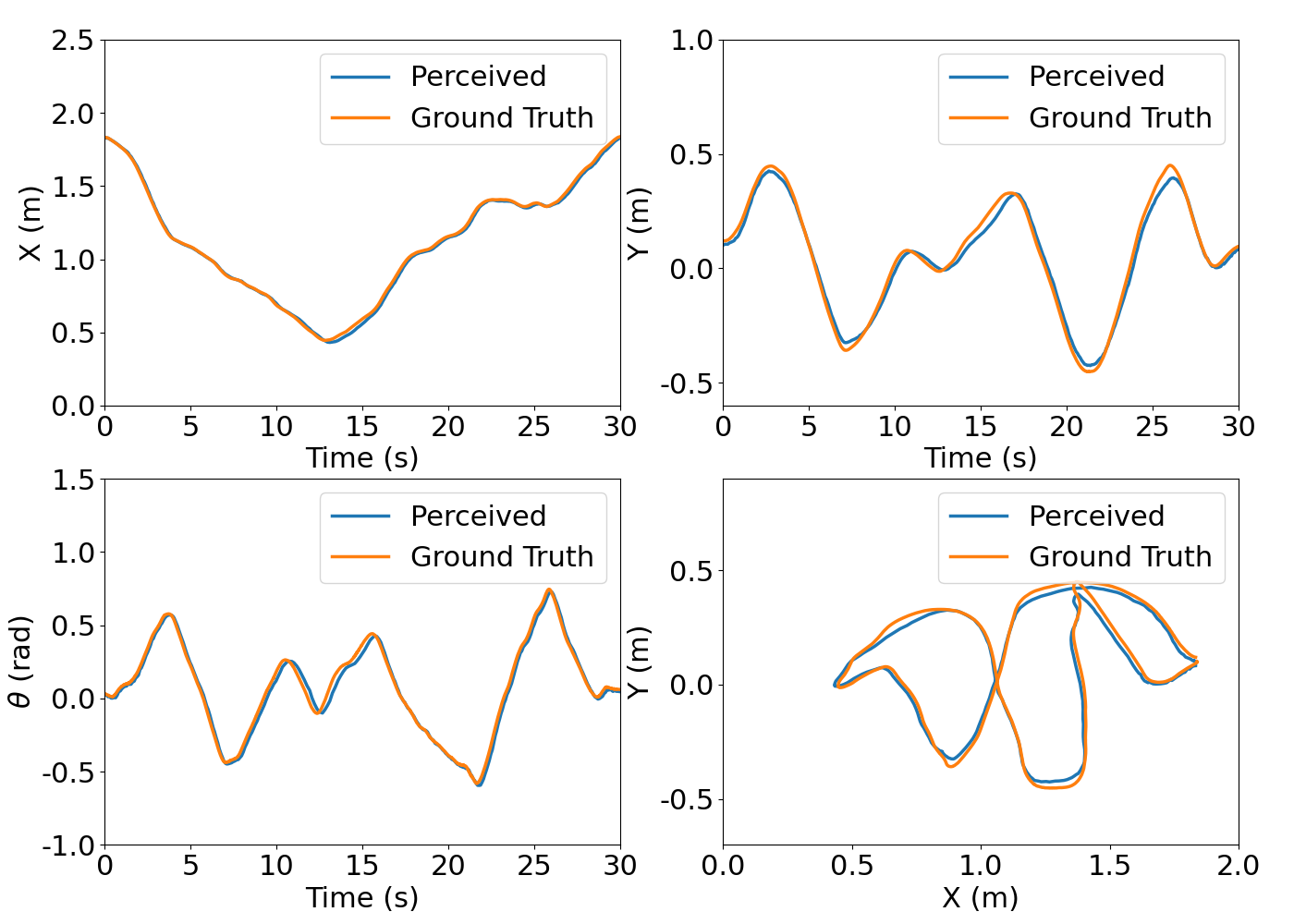}
    \caption{Comparisons between the ground truth pose of a moving trolley and results of our LiDAR-based plane detection method.}
    \label{fig:expPercption2}
    \vspace{-0.5cm}
\end{figure}

To verify the effectiveness of our perception strategy, we conducted experiments moving a trolley in irregular motions and comparing the 3D poses detected by the robot with ground truths measured by a motion capture system. 
First, we tested our 3D monocular method used in long-distance perception, and we
show comparisons between the perceived poses of a moving trolley and the ground truth in \figref{fig:expPercption1}. 
The average estimate error in position is 0.17m with a variance of 0.0097m$^2$, and the average estimated angle error is 0.11rad with a variance of 0.0085rad$^2$. 
Then the proposed short-distance LiDAR method based on plane detection is also validated, and the results are presented in \figref{fig:expPercption2}. 
The average estimate error in position is 0.03m with a variance of 0.0002m$^2$ and the average estimate error in orientation is 0.02rad with a variance of 0.00036rad$^2$.   
In all, the perception module can provide accurate information for further planning and manipulation.

\balance
\subsection{Autonomous Trolley Collection Demonstration}
\label{sub:demo}
\begin{figure}[htbp]
    \centering
    \includegraphics[width=1.0\linewidth]{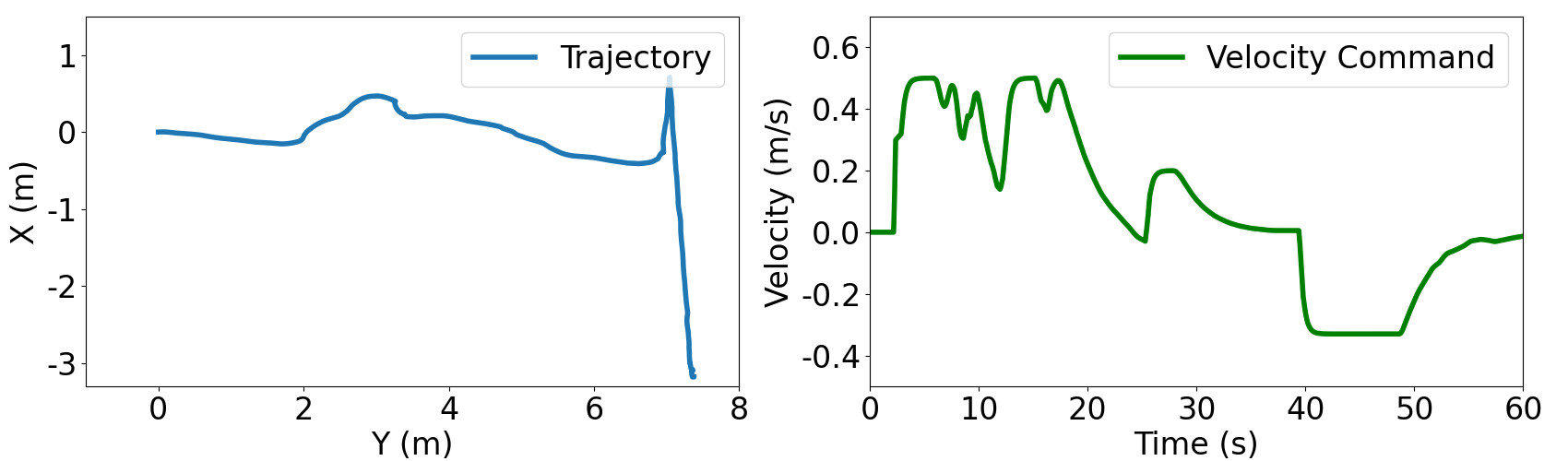}
    \caption{The position trajectory of the robot and the velocity commands yielded by its motion planner.}
    \label{fig:exp1}
    \vspace{-0.4cm}
\end{figure}
Using the approaches described throughout this paper, we demonstrate our system in an actual autonomous trolley collection task. 
The robot is supposed to detect and localize a target trolley, safely navigate itself to the trolley's back, catch the trolley with its manipulator, and finally carry it to a designated returning spot. 
Our hardware setup is shown in \figref{fig:design}, and all the algorithms above were integrated with the Robot Operating System (ROS) environment and run in real time on the robot's onboard computer with an i7-1165G7 CPU and an NVIDIA GTX 2060 GPU. 
In the demonstration shown in \figref{fig:demo}, we put the target trolley at different locations with different orientations, far from several initial locations of the robot, and let the robot perform the collection autonomously. 
In the space between the robot and the target trolley, we set up multiple static obstacles to block the robot's direct route to the goal. 
\figref{fig:exp1} shows the position trajectory of the robot in the demonstration and velocity commands produced by our planner over time. 
In the velocity commands plot in \figref{fig:exp1}, the first big crest happens between $t=10$s and $t=20$s is caused by avoiding the moving human in \figref{subfig:demo1}; the second crest at $t=27$s means the robot has passed the approaching stage and begins docking (see \figref{subfig:demo3}); and the sudden change at $t=39$s indicates the start of the return stage shown in \figref{subfig:demo4}.


\section{Conclusions and Future Work}
\label{sec:conclusion}

In this paper, we propose a mobile manipulation system for robotic autonomous trolley collection in complex and dynamic environments. 
To detect target trolleys and estimate their poses, the robot uses a learning-based 3D detection method involving object and key points detection at long distances, and adopts an accurate point cloud plane detection method at short distances. 
For safe motion planning and control, we model this real-time task as an NMPC problem. 
With CBFs, the obstacle avoidance and field-of-view maintaining requirements are composed into the planning framework as constraints. 
The incorporation of the novel design of mechanical system and autonomy framework together with the progressive perception and planning strategy forms an efficient and robust robotic solution to autonomous trolley collection. 
We demonstrate our system in hardware on an actual trolley collection task with static obstacles and moving humans. 
Experimental results reveal that our solution clearly outperforms most state of the arts regarding the collection task. 
Our future work will focus on developing global decision-making strategies and multi-robot collaboration.


{
    \balance
    \bibliographystyle{IEEEtran}
    \bibliography{IEEEabrv, bib/bibliography}
}

\end{document}